\documentclass[]{ceurart}
\usepackage{graphicx}

\usepackage{wrapfig}
\usepackage{todonotes}
\usepackage{multirow}
\usepackage{rotating}
\usepackage{url}
\usepackage{paralist}
\usepackage{booktabs}

\usepackage{soul}
\usepackage{hyperref}
\usepackage{cleveref}

\makeatletter
\newcommand{\printfnsymbol}[1]{%
  \textsuperscript{\@fnsymbol{#1}}%
}
\makeatother

\begin{document}
\title{Prompting as Probing: Using Language Models for Knowledge Base Construction}
%

\copyrightyear{2022}
\copyrightclause{Copyright for this paper by its authors.
  Use permitted under Creative Commons License Attribution 4.0 International (CC BY 4.0).}

\conference{LM-KBC'22: Knowledge Base Construction from Pre-trained Language Models,
  Challenge at ISWC 2022}

\author[1,2,4]{Dimitrios Alivanistos}[%
email=d.alivanistos@vu.nl,
url=https://dimitrisalivas.github.io/,
]
\author[1,2]{Selene Báez Santamaría}[%
email=s.baezsantamaria@vu.nl,
url=https://selbaez.github.io/,
]
\author[1,2,4]{Michael Cochez}[%
email=m.cochez@vu.nl,
url=https://www.cochez.nl/,
]
\author[1,2,5]{Jan-Christoph Kalo}[%
email=j.c.kalo@vu.nl,
url=https://research.vu.nl/en/persons/jan-christoph-kalo,
]
\author[1,2]{Emile van Krieken}[%
email=e.van.krieken@vu.nl,
url=https://emilevankrieken.com/,
]
\author[1,2, 3]{Thiviyan Thanapalasingam}[%
email=t.thanapalasingam@uva.nl,
url=https://thiviyansingam.com/,
]
\address[1]{Authors are listed in alphabetical order to denote equal contributions.}
\address[2]{Vrije Universiteit Amsterdam}
\address[3]{Universiteit van Amsterdam}
\address[4]{Discovery Lab, Elsevier, The Netherlands}
\address[5]{DReaMS Lab, Huawei, The Netherlands}

\begin{abstract}
Language Models (LMs) have proven to be useful in various downstream applications, such as summarisation, translation, question answering and text classification. LMs are becoming increasingly important tools in Artificial Intelligence, because of the vast quantity of information they can store. In this work, we present ProP (Prompting as Probing), which utilizes GPT-3, a large Language Model originally proposed by OpenAI in 2020, to perform the task of Knowledge Base Construction (KBC). ProP implements a multi-step approach that combines a variety of prompting techniques to achieve this. Our results show that 
manual prompt curation is essential, 
that the LM must be encouraged to give answer sets of variable lengths, in particular including empty answer sets, 
that true/false questions are a useful device to increase precision on suggestions generated by the LM, 
that the size of the LM is a crucial factor,
and that a dictionary of entity aliases improves the LM score. 
Our evaluation study indicates that these proposed techniques can substantially enhance the quality of the final predictions: ProP won track 2 of the LM-KBC competition, outperforming the baseline by 36.4 percentage points. Our implementation is available on \href{https://github.com/HEmile/iswc-challenge}{https://github.com/HEmile/iswc-challenge}.

\end{abstract}

\maketitle
%

%
%
\section{Introduction}
Language Models (LMs) have been at the center of attention, presented as a recent success story of Artificial Intelligence. LMs have shown great promise across a wide range of domains in a variety of different tasks, such as Text Classification \cite{daza2021inductive}, Financial Sentiment Analysis \cite{araci2019finbert}, and Protein Binding Site Prediction \cite{elnaggar2020prottrans}). 
In recent years, prompt engineering for LMs has become a research field in itself, with a plethora of papers working on LM understanding (\emph{e.g.} \cite{prompt-engineering}). 

Natural Language Processing (NLP) researchers have recently investigated whether LMs could potentially be used as Knowledge Bases, by querying for particular information. In Petroni \emph{et al.} \cite{petroni2019language}, the LAMA dataset for probing relational facts from Wikidata in LMs was presented. The authors show that the masked LM BERT can complete Wikidata facts with a precision of around 32\%. Several follow-up papers have pushed this number to almost 50\%~\cite{zhang2022opt,qin-eisner-2021-learning}. While the prediction quality on LAMA is promising, others have argued that LMs should not be used as knowledge graphs, but rather to support the augmentation and curation of KGs~\cite{razniewski2021language}.

In this paper, we describe ProP, the system we implemented for the ``Knowledge Base Construction from Pre-trained Language Models''  challenge at ISWC 2022\footnote{LM-KBC, \url{https://lm-kbc.github.io/}. This work is a submission in the open track (Track 2) in which LMs of any sizes can be used}. The task is to predict possible objects of a triple, where the subject and relation are given. For example, given the string "Ronnie James Dio" as the subject and the relation \textit{PersonInstrument}, an LM needs to predict the answers "bass guitar" or "guitar", and "trumpet", as the objects of the triple. In contrast to the LAMA dataset \cite{petroni2019language}), the LM-KB challenge dataset contains questions for which there are no answers or where multiple answers are valid. Bakel \emph{et al.} argue for precision and recall metrics over the common ranking-based metrics for query answering. \cite{bakel2020approximate}. Since the LM-KBC dataset requires predicting a set of answers and in some cases even empty answer sets, we choose precision and recall metrics to evaluate our system.


ProP uses the GPT-3 model, a large LM proposed by OpenAI in 2020\cite{brown2020language}. 
For each relation type, we engineered prompts that, when given to the language model, probe it to respond with the set of objects for that relation. The components of ProP can be divided into two categories: 
\begin{inparadesc}
\item[prompt generation] focuses on generating the correct prompts that yield the desired answers for the questions in the LM-KBC dataset, and
\item[post-processing] aims to enhance the quality of the predictions.
\end{inparadesc}

\section{Related Work}
\label{sec:related_work}

The development of large pre-trained Language Models (LMs) has led to substantial improvements in NLP research.
It was shown that extensive pre-training on large text corpora encodes large amounts of linguistic and factual knowledge into a language model that can help to improve the performance on various downstream tasks (see \cite{language-models-CACM-overview} for a recent overview).

\textbf{LM as KG}
Petroni et al. \cite{petroni2019language}, and later others \cite{razniewski2021language}, asked the question to what extent language models can replace or support the creation and curation of knowledge graphs.
Petroni et al. proposed the LAMA dataset for probing relational knowledge in language models by using masked language models to complete cloze-style 
sentences. As an example, the language model BERT can complete the sentence ``Paris is the capital of [MASK]" with the word ``France". In this case, it is assumed that the model \textit{knows} about, or can predict, the triple (Paris, capitalOf, France).

While the original paper relied on manually designed prompts for probing the language model, various follow-up works have shown the superiority of automatically learning prompts. Methods can mine prompts from large text corpora and pick the best prompts by applying them to a training dataset as demonstrated in~\cite{jiang2020how,bouraoui2020inducing}.
Prompts can also be directly learned via backpropagation:
BERTESE~\cite{haviv-etal-2021-bertese} and AutoPrompt~\cite{shin2020autoprompt} show how prompts can be learned to improve the performance on LAMA.
The probing performance can be  pushed even further by either directly learning continuous embeddings for the prompts~\cite{qin-eisner-2021-learning,zhong2021factual} or by directly fine-tuning the LM on the training data~\cite{fichtel2021prompt}.
Similar to our work, in FewShot-LAMA~\cite{he2021an} few-shot learning on the original LAMA dataset is evaluated. The authors show that a combination of few-shot examples with learned prompts achieves the best probing results.

Since the publication of the LAMA dataset, a large variety of other probing datasets for factual knowledge in LMs have been created. LAMA-UHN is a more difficult version of LAMA~\cite{poerner2019bert}.  TimeLAMA adds a time component to facts that the model is probed for~\cite{dhingra2021time}. Furthermore, BioLAMA \cite{sung2021can} and MedLAMA \cite{meng2021rewire} are domain-specific probing datasets for biological and medical knowledge.

Most existing approaches have in common that they only probe the language models for entities with a label consisting of a single token from the language model's vocabulary.
Thus, the prediction of long, complex entity names is mostly unexplored.
Furthermore, most existing works have only asked language models to complete a triple with a single prediction, even though some triples might actually allow for multiple possible predictions.
Both these aspects substantially change the challenge of knowledge graph construction.

\textbf{LM for KG}
While probing language models has been heavily studied in the NLP community, the idea of using language models to support knowledge graph curation is not sufficiently studied~\cite{razniewski2021language}. Some works have shown how LMs in combination with knowledge graphs can be used to complete query results~\cite{kalo2020b}. Other works have looked into how to use language models to identify errors in knowledge graphs~\cite{arnaout2022utilizing}, or have studied how to weight KG triples from ConceptNet with language models to measure semantic similarity. Biswas et al.~\cite{biswas2021do} have shown that language models can be used to perform entity typing by predicting the class using language models.

KG-BERT is a system that is most similar to what is required for the KBC Challenge~\cite{yao2019kgbert}. 
A standard BERT model is trained on serialized knowledge graph data to perform link prediction on standard link prediction benchmark datasets. KG-BERT's performance on link prediction is comparable to many state-of-the-art systems that use knowledge graph embeddings for this task.

\textbf{Similar tasks}
This KBC Challenge task is similar to Factual Q\&A with Language models, where the goal is to respond to questions that fall outside the scope of a knowledge base. This shared task differs in that the responses need to include 0 to k answers. Moreover, in the shared task, the factual questions are generated from triples, thus including variation in how a triple might be phrased as a question.

\section{The LM-KBC Challenge}

The LM-KBC dataset contains triples of the form $(s, p, O)$, where $s$ is the textual representation of the \textit{subject}, $p$ is one of 12 different predicates and $O$ the (possibly empty) set of textual representations of \textit{object} entities for prediction. The subjects and objects are of various types. After learning from a training set of such triples, given a new subject and one of the known relations, the task is to predict the complete set of objects. 

\subsection{The LM-KBC Dataset}
\label{sec:data}
For each of the 12 relations, the number of unique subject-entities in the train, dev, and test sets are 100, 50, and 50 respectively. We include detailed distributions of the cardinality (the number of object-entities) for each relation type (Appendix, \Cref{fig:train_set_stats,fig:dev_set_stats}). \Cref{tab:answer_count_averages} in the Appendix shows the aggregated statistics about the number of object-entities, as well as the number of alternative labels per object-entities in the development set. Certain relation types have a much higher average cardinality (e.g. \textit{PersonProfession}=7.42 or \textit{StateSharesBorderState}=5.6) than others (e.g. \textit{CompanyParentOrganization}=0.32, \textit{PersonPlaceOfDeath}=0.50). We also note that only five of the relations allow for empty answer sets. For example, relations associated with a person's death (\textit{PersonPlaceOfDeath} and \textit{PersonCauseOfDeath}) often contain empty answer sets, because many persons in the dataset are still alive.  In these cases, a models needs to be able to predict empty answer sets correctly.

\subsection{The Baseline}
The baseline model is a masking-based approach that uses the popular BERT model \cite{devlin2018bert} in 3 variants (\emph{base-cased}, \emph{large-cased} and \emph{RoBERTa} \cite{liu2019roberta}). The BERT model is tasked with doing prompt completion to predict object-entities for a given subject-relation pair. The prompts used by the baseline approach have been customised for the different relation types.

\section{Our Method}
Previous studies have shown that prompt formulation plays a critical role in the performance of LMs on downstream applications \cite{reynolds2021prompt,prompt-engineering} and this also applies to our work. We investigated different prompting approaches using few-shot learning with GPT-3\ \cite{brown2020language} using OpenAI’s API \footnote{\url{https://openai.com/api} (temperature=0, max\_tokens=100, top\_p=1, frequency\_penalty=0, presence\_penalty=0, logprobs=1)}. In this Section, we first describe how we generate the prompts (prompt generation phase), and followed by how we utilise different components in our pipeline to further fine-tune the prompts for an enhanced performance (post-processing phase).

\subsection{Prompt Generation}

For each relation in the dataset, we manually curate a set of prompt templates consisting of four representative examples\footnote{This is an arbitrary number of training examples. Since few-shot learners are efficient at learning from a handful of examples \cite{brown2020language}, including an excessive number of training examples may not necessarily lead to any improved precision or recall. Therefore, we did not study the effects of varying the number of training examples in the prompts.} selected from the training split. We use these relation-specific templates to generate prompts for every subject entity by replacing a placeholder token in the final line of the template with the subject entity of interest. We task GPT-3 to complete the prompts, and evaluate the \emph{completions} and compute the macro-precision, macro-accuracy and macro-F1 score for each relation.

We ensured that the training examples for few-shot learning included all of the following: (i) questions with answer sets of variable lengths to inform the LM that it can generate multiple answers;
(ii) questions with empty answer sets to ensure that the LM returns nothing when there is no correct answer to a given question; 
(iii) a fixed question-answer order,  where we provide the question and then immediately the answer so that the LM learns the style of the format, and 
(iv) the answer set formatted as a list to ensure we can efficiently parse the answer set from the LM. We did not study the order of these examples, but hypothesize that this is not hugely important to GPT-3 as it can handle long-range dependencies well. 

We formulate the questions either in natural language or in the form of a triple. 
We hand-designed the natural language questions and did not compare them in a structured manner to alternatives. However, we tried out several variations in OpenAI GPT-3 playground to get an intuition of what style of questions are effective. We found those are usually shorter and simpler. 
We include the prompt templates used in Section \ref{prompt-appendix} of the Appendix. In our work, we investigate the use of both prompting styles and compare natural language prompts with triple-based prompts for the different relations. 

\subsection{Empty Answer sets}

There are some questions in the dataset for which the correct answer is an empty answer set. For instance, there are no countries that share borders with Fiji. The empty answer set can be represented as either an empty list \verb|[]|, or as a string within a list \verb|[`None']|. We have observed (see Table \ref{table:results_per_rel_davinci_empty}) that the way that empty sets are represented affects the precision and recall of our approach. Allowing the explicit answer `None' encourages the LM to exclude answers that it is uncertain about.

\subsection{Fact probing}
Our initial results indicated that the recall of our approach was high, but the precision for certain relations was low. 
Therefore, we add a post-processing step called \emph{fact probing} in ProP's pipeline. We use fact probing to ask the LM to probe whether each completion proposed by the LM in the previous step is correct. Inspired by maieutic prompting \cite{jung2022maieutic}, we create a simple prompt, where we translate each predicted completion into a natural language fact. Then we ask the LM to predict whether this fact is true or false. One example of a fact-probing prompt is \textit{Niger neighbours Libya TRUE} for the \textit{CountryBordersWithCountry} relation. We ensure that the LM only predicts either \verb|TRUE| or \verb|FALSE| by adding a true and a false example to the prompt.


\section{Results}
In this Section, we analyse the ProP pipeline we built for generating prompts and evaluate the contribution of each component. We explain how we combine the best-performing components to yield a prediction that obtains a high macro F1-score on the test split. 

\subsection{Prompt-Fine Tuning}

\subsubsection{Natural Language vs Triple-based Prompts}
Table \ref{table:results_per_rel_davinci_triple} shows the quality of the predictions for the natural language prompts and triple-based prompts. 
We note that on F1, the performance between these two prompt styles is mixed, with F1 being higher on the triple-style prompts in only seven out of twelve cases. Unpacking the F1 into recall and precision shows that the triple-style prompts yield higher precision, while natural language prompts yield higher recall. Overall, the triple-style prompts do yield a higher F1 when averaged over each relation. Our intuition is that natural language prompts contain certain words that badly influence the precision of the predictions. It is, however, difficult to study this systematically as the enumeration of word combinations in the prompts is very large. Triple-based prompts circumvent this problem because they only contain the relevant terms in the prompts for the subject entities and relations that are required to predict the object entities.

\begin{table}[h]
\centering
\scriptsize{
\caption{
\label{table:results_per_rel_davinci_triple} Precision, Recall, and F1-score for predictions generated using natural language and triple-based prompts across the different relations. Results are on the dev dataset. Values rounded up to three decimal places. Best scores are in \textbf{bold}.}
\begin{tabular}{p{4.5cm}|p{1cm}p{1.2cm}|p{1cm}p{1.2cm}|p{1cm}p{1.2cm}}
\toprule
\textbf{Relation Type}  & \multicolumn{2}{c|}{\textbf{Precision}} & \multicolumn{2}{c|}{\textbf{Recall}} & \multicolumn{2}{c}{\textbf{F1}} \\ 
\midrule
\textbf{Method}  & Triple     & Natural Language  & Triple     & Natural Language  & Triple     & Natural Language  \\ 
\midrule
ChemicalCompoundElement   & \textbf{0.976} & 0.895 & \textbf{0.919} & 0.885 & \textbf{0.940} & 0.884 \\
CompanyParentOrganization & \textbf{0.587} & 0.385 & \textbf{0.600} & 0.400 & \textbf{0.590} & 0.388 \\ 
CountryBordersWithCountry & \textbf{0.865} & 0.809 & 0.733 & \textbf{0.800} & 0.766 & \textbf{0.785} \\ 
CountryOfficialLanguage   & \textbf{0.933} & 0.798 & 0.810 & \textbf{0.882} & \textbf{0.833} & 0.785 \\ 
PersonCauseOfDeath        & \textbf{0.560} & 0.500 & \textbf{0.550} & 0.500 & \textbf{0.553} & 0.500 \\ 
PersonEmployer            & 0.261 & \textbf{0.273} & 0.267 & \textbf{0.323} & 0.226 & \textbf{0.262} \\ 
PersonInstrument          & \textbf{0.547} & 0.489 & \textbf{0.508} & 0.458 & \textbf{0.502} & 0.446 \\ 
PersonLanguage            & \textbf{0.840} & 0.750 & 0.894 & \textbf{0.932} & \textbf{0.827} & 0.793 \\ 
PersonPlaceOfDeath        & 0.820 & \textbf{0.840} & 0.820 & \textbf{0.840} & 0.820 & \textbf{0.840} \\ 
PersonProfession          & 0.669 & \textbf{0.713} & 0.527 & \textbf{0.535} & 0.556 & \textbf{0.581} \\ 
RiverBasinsCountry        & \textbf{0.845} & 0.820 & \textbf{0.868} & 0.863 & \textbf{0.832} & 0.822 \\ 
StateSharesBorderState    & 0.587 & 0.\textbf{628} & 0.407 & \textbf{0.462} & 0.472 & \textbf{0.522} \\ 
\midrule
\textbf{Average over all relations} & \textbf{0.707} & 0.658 & \textbf{0.658} & 0.657 & \textbf{0.660} & 0.634 \\ \bottomrule
\end{tabular}}
\end{table}

\subsubsection{Empty vs None}
As explained in Section \ref{sec:data}, five out of twelve relations allow for empty answer sets. We experiment with the different ways to represent such empty answer sets, and Table \ref{table:results_per_rel_davinci_empty} shows the results. Three relations get a performance boost when prompted with `NONE' (\textit{CompanyParentOrganization}, \textit{PersonCauseOfDeath}, \textit{PersonInstrument}), while the other two relations perform better when using empty lists (\textit{CountryBordersWithCountry}, \textit{PersonPlaceOfDeath}). For subsequent experiments, we modified the prompt of each relation to use the best performant representation. 

\begin{table}[h]
\centering
\scriptsize{
\caption{
\label{table:results_per_rel_davinci_empty} Precision, recall, and F1-score for the Davinci model across the different relations. Results are on the dev dataset. We only include those relations which require empty answer sets. Values are rounded up to three decimal places. Best scores are in \textbf{bold}.}
\begin{tabular}{p{5cm}|p{1cm}p{1cm}|p{1cm}p{1cm}|p{1cm}p{1cm}}
\toprule
\textbf{Relation Type}  & \multicolumn{2}{c|}{\textbf{Precision}} & \multicolumn{2}{|c|}{\textbf{Recall}} & \multicolumn{2}{|c}{\textbf{F1}} \\ 
\midrule
\textbf{Method}  & Empty    & None  & Empty     & None  & Empty     & None  \\ 
\midrule
CompanyParentOrganization & 0.587 & \textbf{0.767} & 0.600 & \textbf{0.780} & 0.590 & \textbf{0.770} \\ 
CountryBordersWithCountry & \textbf{0.865} & 0.826 & \textbf{0.733} & 0.719 & \textbf{0.766} & 0.749 \\ 
PersonCauseOfDeath        & 0.560 & \textbf{0.600} & 0.550 & \textbf{0.590} & 0.553 & \textbf{0.593} \\ 
PersonInstrument          & 0.547 & \textbf{0.600} & 0.508 & \textbf{0.561} & 0.502 & \textbf{0.568} \\ 
PersonPlaceOfDeath        & \textbf{0.820} & 0.780 & \textbf{0.820} & 0.780 & \textbf{0.820} & 0.780 \\ 
\midrule
\textbf{Average over all relations} & 0.685 & \textbf{0.697} & 0.674 & \textbf{0.685} & 0.657 & \textbf{0.669} \\ 
\bottomrule
\end{tabular}}
\end{table}

\subsubsection{Language Model Size}
\begin{figure*}[ht!]
    \centering
    \includegraphics[width=.8\linewidth]{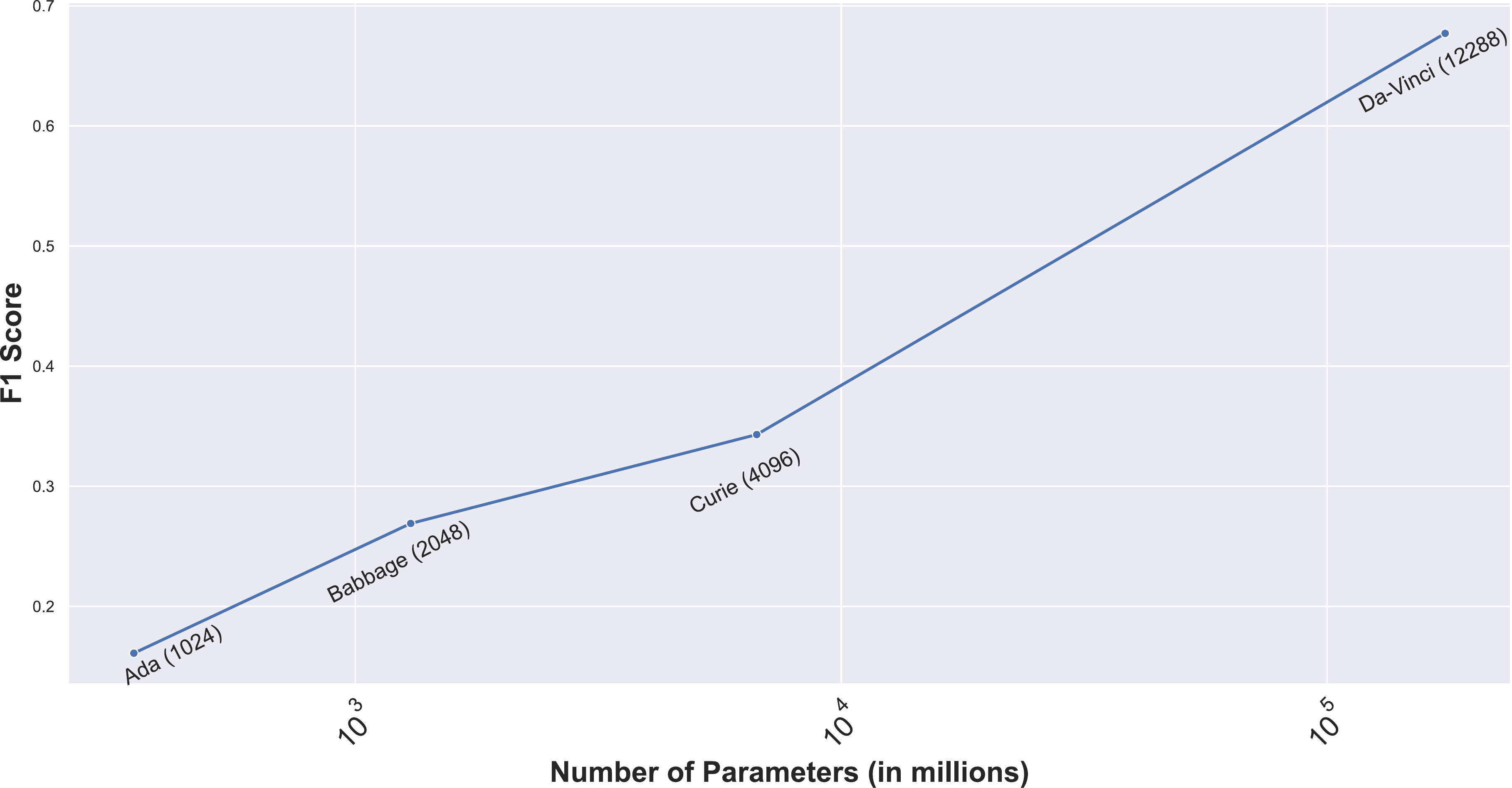}
    \caption{\label{fig:performance_per_model_type}
    The F1-scores of GPT-3 models with different number of parameters. In brackets, the embedding dimension for the model. We observe an almost analogous increase between size and performance.}
\end{figure*}

Usually, the size of LMs is measured by the number of learnable parameters in the model. However, the OpenAI API does not quantify the number of total parameters but only the size of the embedding dimensions for the tokens \footnote{https://beta.openai.com/docs/models/gpt-3}. We assume there is a positive correlation between the token dimension size and the total number of GPT-3 parameters. Figure \ref{fig:performance_per_model_type} shows our results, and we can see that as the language model size increases, the F1-score also increases. This shows that a larger LM gives better performance on KBC.

\subsection{Post-Processing Predictions}

Up to this point, we have discussed how we generate the optimal prompts for the different relation types. Once the LM produces the completions using these optimal prompt techniques, we can employ two additional steps to enhance the precision and recall of our predictions. Table \ref{table:results-probing} shows the results of including fact probing and entity aliases in our system. 

\subsubsection{Fact probing}
We found that fact probing has a different impact on different relation types. This difference could stem from the cardinalities of the relation types. For example, the relation \textit{PersonPlaceOfDeath}, which has only one correct answer, should show a larger improvement than \textit{State Borders}, which has a higher cardinality. We found that fact probing helped to boost the predictions of five relations (\emph{CompanyParentOrganization, CountryOfficialLanguage, PersonCauseOfDeath, PersonInstrument, PersonLanguage}). We only apply fact probing to these relations. On the dev set, the precision of fact probing is 0.737, and 0.608 among the predictions removed by fact probing. That is, in 60.8\% fact probing filtered a prediction, it correctly removed a prediction that was not in the ground truth set.

\begin{table*}[h]
\centering
\scriptsize{
\caption{
\label{table:results-probing}Precision, recall, and F1-score for predictions generated using different post-processing techniques on the development (dev) and test sets. We round up values to three decimal places. Best F1-scores are in \textbf{bold}. }
\begin{tabular}{p{6cm}|p{1.5cm}p{1.5cm}p{1.5cm}}
\toprule
\textbf{Method} & \textbf{Precision} & \textbf{Recall} & \textbf{F1} \\ 
\midrule
BERT Baseline (dev) & 0.349  & 0.295  & 0.309 \\ 
BERT Baseline + fact probing (dev) &  0.357 & 0.304 & \textbf{0.317} \\ 
BERT Baseline + fact probing + alias (dev) & 0.357 & 0.304 & 0.317  \\
\midrule
GPT3-Davinci (dev) & 0.736 &	0.699 &	0.697 \\ 
GPT3-Davinci +  fact probing (dev) & 0.741 &	0.692 &	0.698 \\ 
GPT3-Davinci +  fact probing + alias (dev) &  0.755 &	0.709 &	\textbf{0.712} \\ 
\midrule
GPT3-Davinci (test) & 0.782 &	0.701 &	0.679 \\ 
GPT3-Davinci +  fact probing (test) & 0.798 &	0.690 &	0.676 \\ 
GPT3-Davinci +  fact probing + alias (test) & 0.813	& 0.704	& \textbf{0.689} \\ 
\bottomrule
\end{tabular}}
\end{table*}


\subsubsection{Entity aliases}

As we discussed in \Cref{sec:data}, the predictions from the language model are sometimes correct, but not according to what the gold standard expects.
Whether this is problematic depends on the final use of the predictive model. 
In interactive use, this would not be an issue because the user will be able to disambiguate.
For actual KBC, the system will have to disambiguate what exact entity it predicted. 
Here, however, we only check whether the text generated by the model corresponds to one of the gold standard alternatives.

While experimenting, we noticed that in the training and development datasets the names of entities often correspond with the labels of entities in Wikidata\footnote{\url{https://www.wikidata.org}}. 
On Wikidata, these entities also have aliases, and we wanted to know whether we can improve our system by looking up the aliases on Wikidata. This lookup does not use language models, so it is \emph{not included as part of the ProP pipeline}, as this would violate the terms of the LM-KBC challenge. 
Instead, we perform it as an ablation study.

The \textit{alias-fetcher} works as follows. First, we extract a set of types which could be relevant for the specific relation types. For example, country (Q6256) is relevant for \textit{RiverBasinsCountry}. Then, for each relation type, we extract all correctly typed entities, their aliases, and claim count.
Then, we take the prediction of the LM, and check whether there is an entity with that label for that relation. If so, we retain the prediction.

Otherwise, we check whether the prediction is equal to any alias. There could be multiple entities for which this is the case. Therefore, we pick the label of the entity with the most claims on Wikidata. The assumption is that it is more likely the answer if it is an `interesting' entity and that these have more claims on Wikidata.

We observe that for four relation types, the changes in the scores are insignificant. For the eight other relations, we see that the F1 score goes up slightly. Overall, this results in an average improvement of the F1 score with 0.014 (Table \ref{table:results-probing}) on the development set. On the test set, we notice a similar improvement. This experiment is not extensive enough to derive definitive conclusions, but it appears to be useful to use structured data to augment the predictions of an LM.

\subsubsection{Contemporaneity of LMs}

We found that questions regarding recent events, particularly those that occurred after 2020, did not yield good predictions by GPT-3 (see \ref{sec:failure-cases} in the Appendix). 
This is in line with related findings around LMs and was confirmed by OpenAI \footnote{\url{https://beta.openai.com/docs/guides/embeddings/limitations-risks}}. Two examples of this include \emph{Facebook, Inc.} changing its name to \emph{Meta Platforms, Inc.} (in October 2021), and the country of \emph{Swaziland} changing its name to \emph{Eswatini} (in 2018). We also observed similar problems with several instances from the following relations: PersonProfession, PersonCauseOfDeath, and PersonPlaceOfDeath. It is worth noting that it is not important when the model was trained, but whether the training date contains up-to-date information.

\subsection{Future Work}

A natural continuation of this work revolves around improving the individual steps in our pipeline (\emph{e.g.} fact checking) and their performance, which will directly reflect on the overall macro-F1 score of our approach. Additionally, we could experiment with inverting our pipeline and allow the LM to generate the best prompts by providing the ground truth as input.
For example, we could explore techniques that automatically learn prompts, similar to AutoPrompt~\cite{shin2020autoprompt} and OptiPrompt~\cite{zhang2022opt}, but ideally with a method that requires fewer resources.

In terms of additional components that make use of LMs, we considered developing \textit{meta-prompts} as in \cite{reynolds2021prompt}. We think it would be interesting to study what meta-prompts can be developed for KBC. Is there a set of specific patterns that work better than others? 
Finally, we could modify our \textit{alias-fetcher} to use the LM to generate well-known aliases for both entities and relations in the training data. This approach could act as a diversification factor, and we believe it will have more freedom in its choice of aliases.

Data augmentation differs from \textit{prompt tuning} in the following ways: While prompt tuning is looking for the optimal prompt to increase the performance for a specific task, \textit{data augmentation} acts as a diversification mechanism for our existing prompts. By employing a more diverse set of prompts we can increase our performance, especially the recall. We base our hypothesis on the fact that knowledge is expressed diversely in the training data (e.g. ambiguity), and we believe this should be considered when prompting an LM.

Furthermore, it would be interesting to further investigate if huge language models are required to perform knowledge graph construction and how to achieve the best prediction performance for the lowest costs.

\section{Conclusion}
We introduced ProP, our "Prompting as Probing" approach to performing knowledge-base completion using a high-capacity pre-trained LM. We showed how we developed different modular components that utilise both LM and the data provided by the organisers to improve ProP's performance, such as the \textit{fact probing} and the \textit{alias fetcher} components. We also investigated well-known techniques around prompt engineering and optimisation and analysed the effect of different prompt formulations on the final performance. However, we conclude that the parameter count of the GPT-3 models is the most significant contributor to performance. Our ProP pipeline outperforms the baseline by 36.4 percentage points on the test split. 

Our approach does not only obtain a high macro F1-score on the ground truth but its actual score is suspected to be higher because in several cases where the result was counted as incorrect, the ground truth was either incomplete or used aliases that refer to the same entity as our prediction. Overall, we conclude that language models can be used to augment Knowledge Bases, and we emphasise the difficulty of evaluating question-answering tasks where simple string matching does not suffice.


\paragraph*{Supplemental Material Statement:}Code and data are publicly available from
\url{https://github.com/HEmile/iswc-challenge}. 

\subsubsection*{Acknowledgements}
We thank Frank van Harmelen for his insightful comments. This research was funded by the Vrije Universiteit Amsterdam and the Netherlands Organisation for Scientific Research (NWO) via the \textit{Spinoza} grant (SPI 63-260) awarded to Piek Vossen,  the \textit{Hybrid Intelligence Centre} via the Zwaartekracht grant (024.004.022), Elsevier’s Discovery Lab, and Huawei's DReaMS Lab.

%
%
%
\bibliography{mybib}

\begin{thebibliography}{30}
\expandafter\ifx\csname natexlab\endcsname\relax\def\natexlab#1{#1}\fi
\providecommand{\url}[1]{\texttt{#1}}
\providecommand{\href}[2]{#2}
\providecommand{\path}[1]{#1}
\providecommand{\DOIprefix}{doi:}
\providecommand{\ArXivprefix}{arXiv:}
\providecommand{\URLprefix}{URL: }
\providecommand{\Pubmedprefix}{pmid:}
\providecommand{\doi}[1]{\href{http://dx.doi.org/#1}{\path{#1}}}
\providecommand{\Pubmed}[1]{\href{pmid:#1}{\path{#1}}}
\providecommand{\bibinfo}[2]{#2}
\ifx\xfnm\relax \def\xfnm[#1]{\unskip,\space#1}\fi
\bibitem[{Daza et~al.(2021)Daza, Cochez, and Groth}]{daza2021inductive}
\bibinfo{author}{D.~Daza}, \bibinfo{author}{M.~Cochez},
  \bibinfo{author}{P.~Groth},
\newblock \bibinfo{title}{Inductive entity representations from text via link
  prediction},
\newblock in: \bibinfo{booktitle}{Proceedings of the Web Conference 2021},
  \bibinfo{year}{2021}, pp. \bibinfo{pages}{798--808}.
\bibitem[{Araci(2019)}]{araci2019finbert}
\bibinfo{author}{D.~Araci},
\newblock \bibinfo{title}{Finbert: Financial sentiment analysis with
  pre-trained language models},
\newblock \bibinfo{journal}{arXiv preprint arXiv:1908.10063}
  (\bibinfo{year}{2019}).
\bibitem[{Elnaggar et~al.(2020)Elnaggar, Heinzinger, Dallago, Rihawi, Wang
  et~al.}]{elnaggar2020prottrans}
\bibinfo{author}{A.~Elnaggar}, \bibinfo{author}{M.~Heinzinger},
  \bibinfo{author}{C.~Dallago}, \bibinfo{author}{G.~Rihawi},
  \bibinfo{author}{Y.~Wang}, et~al.,
\newblock \bibinfo{title}{{ProtTrans}: Towards cracking the language of life's
  code through self-supervised deep learning and high performance computing},
\newblock \bibinfo{journal}{arXiv preprint arXiv:2007.06225}
  (\bibinfo{year}{2020}).
\bibitem[{Sorensen et~al.(2022)Sorensen, Robinson, Rytting, Shaw, Rogers,
  Delorey, Khalil, Fulda, and Wingate}]{prompt-engineering}
\bibinfo{author}{T.~Sorensen}, \bibinfo{author}{J.~Robinson},
  \bibinfo{author}{C.~M. Rytting}, \bibinfo{author}{A.~G. Shaw},
  \bibinfo{author}{K.~J. Rogers}, \bibinfo{author}{A.~P. Delorey},
  \bibinfo{author}{M.~Khalil}, \bibinfo{author}{N.~Fulda},
  \bibinfo{author}{D.~Wingate}, \bibinfo{title}{An information-theoretic
  approach to prompt engineering without ground truth labels},
  \bibinfo{year}{2022}.
\bibitem[{Petroni et~al.(2019)Petroni, Rockt{\"{a}}schel, Riedel, Lewis,
  Bakhtin, Wu, and Miller}]{petroni2019language}
\bibinfo{author}{F.~Petroni}, \bibinfo{author}{T.~Rockt{\"{a}}schel},
  \bibinfo{author}{S.~Riedel}, \bibinfo{author}{P.~Lewis},
  \bibinfo{author}{A.~Bakhtin}, \bibinfo{author}{Y.~Wu},
  \bibinfo{author}{A.~Miller},
\newblock \bibinfo{title}{{Language Models as Knowledge Bases?}},
\newblock in: \bibinfo{booktitle}{Proc. of the 2019 Conference on Empirical
  Methods in Natural Language Processing and the 9th International Joint
  Conference on Natural Language Processing (EMNLP-IJCNLP)},
  \bibinfo{year}{2019}, pp. \bibinfo{pages}{2463--2473}.
\bibitem[{Zhang et~al.(2022)Zhang, Roller, Goyal, Artetxe, Chen, Chen, Dewan,
  Diab, Li, Lin, Mihaylov, Ott, Shleifer, Shuster, Simig, Koura, Sridhar, Wang,
  and Zettlemoyer}]{zhang2022opt}
\bibinfo{author}{S.~Zhang}, \bibinfo{author}{S.~Roller},
  \bibinfo{author}{N.~Goyal}, \bibinfo{author}{M.~Artetxe},
  \bibinfo{author}{M.~Chen}, \bibinfo{author}{S.~Chen},
  \bibinfo{author}{C.~Dewan}, \bibinfo{author}{M.~Diab},
  \bibinfo{author}{X.~Li}, \bibinfo{author}{X.~V. Lin},
  \bibinfo{author}{T.~Mihaylov}, \bibinfo{author}{M.~Ott},
  \bibinfo{author}{S.~Shleifer}, \bibinfo{author}{K.~Shuster},
  \bibinfo{author}{D.~Simig}, \bibinfo{author}{P.~S. Koura},
  \bibinfo{author}{A.~Sridhar}, \bibinfo{author}{T.~Wang},
  \bibinfo{author}{L.~Zettlemoyer}, \bibinfo{title}{Opt: Open pre-trained
  transformer language models}, \bibinfo{year}{2022}.
  \href{http://arxiv.org/abs/2205.01068}{{\tt arXiv:2205.01068}}.
\bibitem[{Qin and Eisner(2021)}]{qin-eisner-2021-learning}
\bibinfo{author}{G.~Qin}, \bibinfo{author}{J.~Eisner},
\newblock \bibinfo{title}{Learning how to ask: Querying {LM}s with mixtures of
  soft prompts},
\newblock in: \bibinfo{booktitle}{Proceedings of the 2021 Conference of the
  North American Chapter of the Association for Computational Linguistics:
  Human Language Technologies}, \bibinfo{year}{2021}, pp.
  \bibinfo{pages}{5203--5212}.
\bibitem[{Razniewski et~al.(2021)Razniewski, Yates, Kassner, and
  Weikum}]{razniewski2021language}
\bibinfo{author}{S.~Razniewski}, \bibinfo{author}{A.~Yates},
  \bibinfo{author}{N.~Kassner}, \bibinfo{author}{G.~Weikum},
\newblock \bibinfo{title}{{Language Models As or For Knowledge Bases}}
  (\bibinfo{year}{2021}). \href{http://arxiv.org/abs/2110.04888}{{\tt
  arXiv:2110.04888}}.
\bibitem[{Bakel et~al.(2020)Bakel, Aleksiev, Daza, Alivanistos, and
  Cochez}]{bakel2020approximate}
\bibinfo{author}{R.~v. Bakel}, \bibinfo{author}{T.~Aleksiev},
  \bibinfo{author}{D.~Daza}, \bibinfo{author}{D.~Alivanistos},
  \bibinfo{author}{M.~Cochez},
\newblock \bibinfo{title}{Approximate knowledge graph query answering: from
  ranking to binary classification},
\newblock in: \bibinfo{booktitle}{International Workshop on Graph Structures
  for Knowledge Representation and Reasoning}, \bibinfo{organization}{Springer,
  Cham}, \bibinfo{year}{2020}, pp. \bibinfo{pages}{107--124}.
\bibitem[{Brown et~al.(2020)Brown, Mann, Ryder, Subbiah, Kaplan, Dhariwal,
  Neelakantan, Shyam, Sastry, Askell et~al.}]{brown2020language}
\bibinfo{author}{T.~Brown}, \bibinfo{author}{B.~Mann},
  \bibinfo{author}{N.~Ryder}, \bibinfo{author}{M.~Subbiah},
  \bibinfo{author}{J.~D. Kaplan}, \bibinfo{author}{P.~Dhariwal},
  \bibinfo{author}{A.~Neelakantan}, \bibinfo{author}{P.~Shyam},
  \bibinfo{author}{G.~Sastry}, \bibinfo{author}{A.~Askell}, et~al.,
\newblock \bibinfo{title}{Language models are few-shot learners},
\newblock \bibinfo{journal}{Advances in neural information processing systems}
  \bibinfo{volume}{33} (\bibinfo{year}{2020}) \bibinfo{pages}{1877--1901}.
\bibitem[{Li(2022)}]{language-models-CACM-overview}
\bibinfo{author}{H.~Li},
\newblock \bibinfo{title}{Language models: past, present, and future},
\newblock \bibinfo{journal}{Commun. {ACM}} \bibinfo{volume}{65}
  (\bibinfo{year}{2022}) \bibinfo{pages}{56--63}.
\bibitem[{Jiang et~al.(2020)Jiang, Xu, Araki, and Neubig}]{jiang2020how}
\bibinfo{author}{Z.~Jiang}, \bibinfo{author}{F.~F. Xu},
  \bibinfo{author}{J.~Araki}, \bibinfo{author}{G.~Neubig},
\newblock \bibinfo{title}{{How Can We Know What Language Models Know?}},
\newblock in: \bibinfo{booktitle}{Transactions of the Association for
  Computational Linguistics 2020 (TACL)}, volume~\bibinfo{volume}{8},
  \bibinfo{year}{2020}, pp. \bibinfo{pages}{423--438}.
\bibitem[{Bouraoui et~al.(2020)Bouraoui, Camacho-Collados, and
  Schockaert}]{bouraoui2020inducing}
\bibinfo{author}{Z.~Bouraoui}, \bibinfo{author}{J.~Camacho-Collados},
  \bibinfo{author}{S.~Schockaert},
\newblock \bibinfo{title}{{Inducing Relational Knowledge from BERT}},
\newblock in: \bibinfo{booktitle}{Proc. of the Thirty-Fourth Conference on
  Artificial Intelligence}, AAAI'20, \bibinfo{year}{2020}.
\bibitem[{Haviv et~al.(2021)Haviv, Berant, and
  Globerson}]{haviv-etal-2021-bertese}
\bibinfo{author}{A.~Haviv}, \bibinfo{author}{J.~Berant},
  \bibinfo{author}{A.~Globerson},
\newblock \bibinfo{title}{{BERT}ese: Learning to speak to {BERT}},
\newblock in: \bibinfo{booktitle}{Proceedings of the 16th Conference of the
  European Chapter of the Association for Computational Linguistics: Main
  Volume}, \bibinfo{year}{2021}, pp. \bibinfo{pages}{3618--3623}.
\bibitem[{Shin et~al.(2020)Shin, Razeghi, {Logan IV}, Wallace, and
  Singh}]{shin2020autoprompt}
\bibinfo{author}{T.~Shin}, \bibinfo{author}{Y.~Razeghi}, \bibinfo{author}{R.~L.
  {Logan IV}}, \bibinfo{author}{E.~Wallace}, \bibinfo{author}{S.~Singh},
\newblock \bibinfo{title}{{AutoPrompt: Eliciting Knowledge from Language Models
  with Automatically Generated Prompts}}  (\bibinfo{year}{2020})
  \bibinfo{pages}{4222--4235}. \href{http://arxiv.org/abs/2010.15980}{{\tt
  arXiv:2010.15980}}.
\bibitem[{Zhong et~al.(2021)Zhong, Friedman, and Chen}]{zhong2021factual}
\bibinfo{author}{Z.~Zhong}, \bibinfo{author}{D.~Friedman},
  \bibinfo{author}{D.~Chen},
\newblock \bibinfo{title}{Factual probing is [{MASK}]: Learning vs. learning to
  recall}  (\bibinfo{year}{2021}) \bibinfo{pages}{5017--5033}.
\bibitem[{Fichtel et~al.(2021)Fichtel, Kalo, and Balke}]{fichtel2021prompt}
\bibinfo{author}{L.~Fichtel}, \bibinfo{author}{J.-C. Kalo},
  \bibinfo{author}{W.-T. Balke},
\newblock \bibinfo{title}{{Prompt Tuning or Fine-Tuning - Investigating
  Relational Knowledge in Pre-Trained Language Models}}  (\bibinfo{year}{2021})
  \bibinfo{pages}{1--15}.
\bibitem[{He et~al.(2021)He, Cho, and Glass}]{he2021an}
\bibinfo{author}{T.~He}, \bibinfo{author}{K.~Cho}, \bibinfo{author}{J.~Glass},
\newblock \bibinfo{title}{{An Empirical Study on Few-shot Knowledge Probing for
  Pretrained Language Models}}  (\bibinfo{year}{2021}).
  \href{http://arxiv.org/abs/2109.02772}{{\tt arXiv:2109.02772}}.
\bibitem[{Poerner et~al.(2019)Poerner, Waltinger, and
  Sch{\"{u}}tze}]{poerner2019bert}
\bibinfo{author}{N.~Poerner}, \bibinfo{author}{U.~Waltinger},
  \bibinfo{author}{H.~Sch{\"{u}}tze},
\newblock \bibinfo{title}{{BERT is Not a Knowledge Base (Yet): Factual
  Knowledge vs. Name-Based Reasoning in Unsupervised QA}} \bibinfo{volume}{0}
  (\bibinfo{year}{2019}). \href{http://arxiv.org/abs/1911.03681}{{\tt
  arXiv:1911.03681}}.
\bibitem[{Dhingra et~al.(2021)Dhingra, Cole, Eisenschlos, Gillick, Eisenstein,
  and Cohen}]{dhingra2021time}
\bibinfo{author}{B.~Dhingra}, \bibinfo{author}{J.~R. Cole},
  \bibinfo{author}{J.~M. Eisenschlos}, \bibinfo{author}{D.~Gillick},
  \bibinfo{author}{J.~Eisenstein}, \bibinfo{author}{W.~W. Cohen},
\newblock \bibinfo{title}{{Time-Aware Language Models as Temporal Knowledge
  Bases}}  (\bibinfo{year}{2021}). \href{http://arxiv.org/abs/2106.15110}{{\tt
  arXiv:2106.15110}}.
\bibitem[{Sung et~al.(2021)Sung, Lee, Yi, Jeon, Kim, and Kang}]{sung2021can}
\bibinfo{author}{M.~Sung}, \bibinfo{author}{J.~Lee}, \bibinfo{author}{S.~Yi},
  \bibinfo{author}{M.~Jeon}, \bibinfo{author}{S.~Kim},
  \bibinfo{author}{J.~Kang},
\newblock \bibinfo{title}{{Can Language Models be Biomedical Knowledge Bases?}}
   (\bibinfo{year}{2021}) \bibinfo{pages}{4723--4734}.
  \href{http://arxiv.org/abs/2109.07154}{{\tt arXiv:2109.07154}}.
\bibitem[{Meng et~al.(2021)Meng, Liu, Shareghi, Su, Collins, and
  Collier}]{meng2021rewire}
\bibinfo{author}{Z.~Meng}, \bibinfo{author}{F.~Liu},
  \bibinfo{author}{E.~Shareghi}, \bibinfo{author}{Y.~Su},
  \bibinfo{author}{C.~Collins}, \bibinfo{author}{N.~Collier},
\newblock \bibinfo{title}{{Rewire-then-Probe: A Contrastive Recipe for Probing
  Biomedical Knowledge of Pre-trained Language Models}}
  (\bibinfo{year}{2021}). \href{http://arxiv.org/abs/2110.08173}{{\tt
  arXiv:2110.08173}}.
\bibitem[{Kalo et~al.(2020)Kalo, Fichtel, Ehler, and Balke}]{kalo2020b}
\bibinfo{author}{J.-C. Kalo}, \bibinfo{author}{L.~Fichtel},
  \bibinfo{author}{P.~Ehler}, \bibinfo{author}{W.-T. Balke},
\newblock \bibinfo{title}{{KnowlyBERT - Hybrid Query Answering over Language
  Models and Knowledge Graphs}},
\newblock in: \bibinfo{booktitle}{Proceedings of the International Semantic Web
  Conference (ISWC)}, \bibinfo{year}{2020}, pp. \bibinfo{pages}{294--310}.
\bibitem[{Arnaout et~al.(2022)Arnaout, Tran, Stepanova, Gad-Elrab, Razniewski,
  and Weikum}]{arnaout2022utilizing}
\bibinfo{author}{H.~Arnaout}, \bibinfo{author}{T.-K. Tran},
  \bibinfo{author}{D.~Stepanova}, \bibinfo{author}{M.~H. Gad-Elrab},
  \bibinfo{author}{S.~Razniewski}, \bibinfo{author}{G.~Weikum},
\newblock \bibinfo{title}{Utilizing language model probes for knowledge graph
  repair},
\newblock in: \bibinfo{booktitle}{Wiki Workshop 2022}, \bibinfo{year}{2022}.
\bibitem[{Biswas et~al.(2021)Biswas, Sofronova, Alam, Heist, Paulheim, and
  Sack}]{biswas2021do}
\bibinfo{author}{R.~Biswas}, \bibinfo{author}{R.~Sofronova},
  \bibinfo{author}{M.~Alam}, \bibinfo{author}{N.~Heist},
  \bibinfo{author}{H.~Paulheim}, \bibinfo{author}{H.~Sack},
\newblock \bibinfo{title}{{Do Judge an Entity by Its Name! Entity Typing Using
  Language Models}},
\newblock in: \bibinfo{booktitle}{The Semantic Web: ESWC 2021 Satellite
  Events}, \bibinfo{year}{2021}, pp. \bibinfo{pages}{65--70}.
\bibitem[{Yao et~al.(2019)Yao, Mao, and Luo}]{yao2019kgbert}
\bibinfo{author}{L.~Yao}, \bibinfo{author}{C.~Mao}, \bibinfo{author}{Y.~Luo},
  \bibinfo{title}{Kg-bert: Bert for knowledge graph completion},
  \bibinfo{year}{2019}. \href{http://arxiv.org/abs/1909.03193}{{\tt
  arXiv:1909.03193}}.
\bibitem[{Devlin et~al.(2018)Devlin, Chang, Lee, and
  Toutanova}]{devlin2018bert}
\bibinfo{author}{J.~Devlin}, \bibinfo{author}{M.-W. Chang},
  \bibinfo{author}{K.~Lee}, \bibinfo{author}{K.~Toutanova},
\newblock \bibinfo{title}{Bert: Pre-training of deep bidirectional transformers
  for language understanding},
\newblock \bibinfo{journal}{arXiv preprint arXiv:1810.04805}
  (\bibinfo{year}{2018}).
\bibitem[{Liu et~al.(2019)Liu, Ott, Goyal, Du, Joshi, Chen, Levy, Lewis,
  Zettlemoyer, and Stoyanov}]{liu2019roberta}
\bibinfo{author}{Y.~Liu}, \bibinfo{author}{M.~Ott}, \bibinfo{author}{N.~Goyal},
  \bibinfo{author}{J.~Du}, \bibinfo{author}{M.~Joshi},
  \bibinfo{author}{D.~Chen}, \bibinfo{author}{O.~Levy},
  \bibinfo{author}{M.~Lewis}, \bibinfo{author}{L.~Zettlemoyer},
  \bibinfo{author}{V.~Stoyanov},
\newblock \bibinfo{title}{Roberta: A robustly optimized bert pretraining
  approach},
\newblock \bibinfo{journal}{arXiv preprint arXiv:1907.11692}
  (\bibinfo{year}{2019}).
\bibitem[{Reynolds and McDonell(2021)}]{reynolds2021prompt}
\bibinfo{author}{L.~Reynolds}, \bibinfo{author}{K.~McDonell},
\newblock \bibinfo{title}{Prompt programming for large language models: Beyond
  the few-shot paradigm},
\newblock in: \bibinfo{booktitle}{Extended Abstracts of the 2021 CHI Conference
  on Human Factors in Computing Systems}, \bibinfo{year}{2021}, pp.
  \bibinfo{pages}{1--7}.
\bibitem[{Jung et~al.(2022)Jung, Qin, Welleck, Brahman, Bhagavatula, Bras, and
  Choi}]{jung2022maieutic}
\bibinfo{author}{J.~Jung}, \bibinfo{author}{L.~Qin},
  \bibinfo{author}{S.~Welleck}, \bibinfo{author}{F.~Brahman},
  \bibinfo{author}{C.~Bhagavatula}, \bibinfo{author}{R.~L. Bras},
  \bibinfo{author}{Y.~Choi},
\newblock \bibinfo{title}{Maieutic prompting: Logically consistent reasoning
  with recursive explanations},
\newblock \bibinfo{journal}{arXiv preprint arXiv:2205.11822}
  (\bibinfo{year}{2022}).

\end{thebibliography}
\newpage
\section{Appendix}

\subsection{Dataset statistics}

Here, we provide statistics about the LM-KBC dataset for the training and development split. The statistics of the test split is unknown, because the test split is not public. We assume that the instances from the test are also sampled from a similar data distribution.

\begin{figure}[ht]
    \centering
    \includegraphics[width=\linewidth]{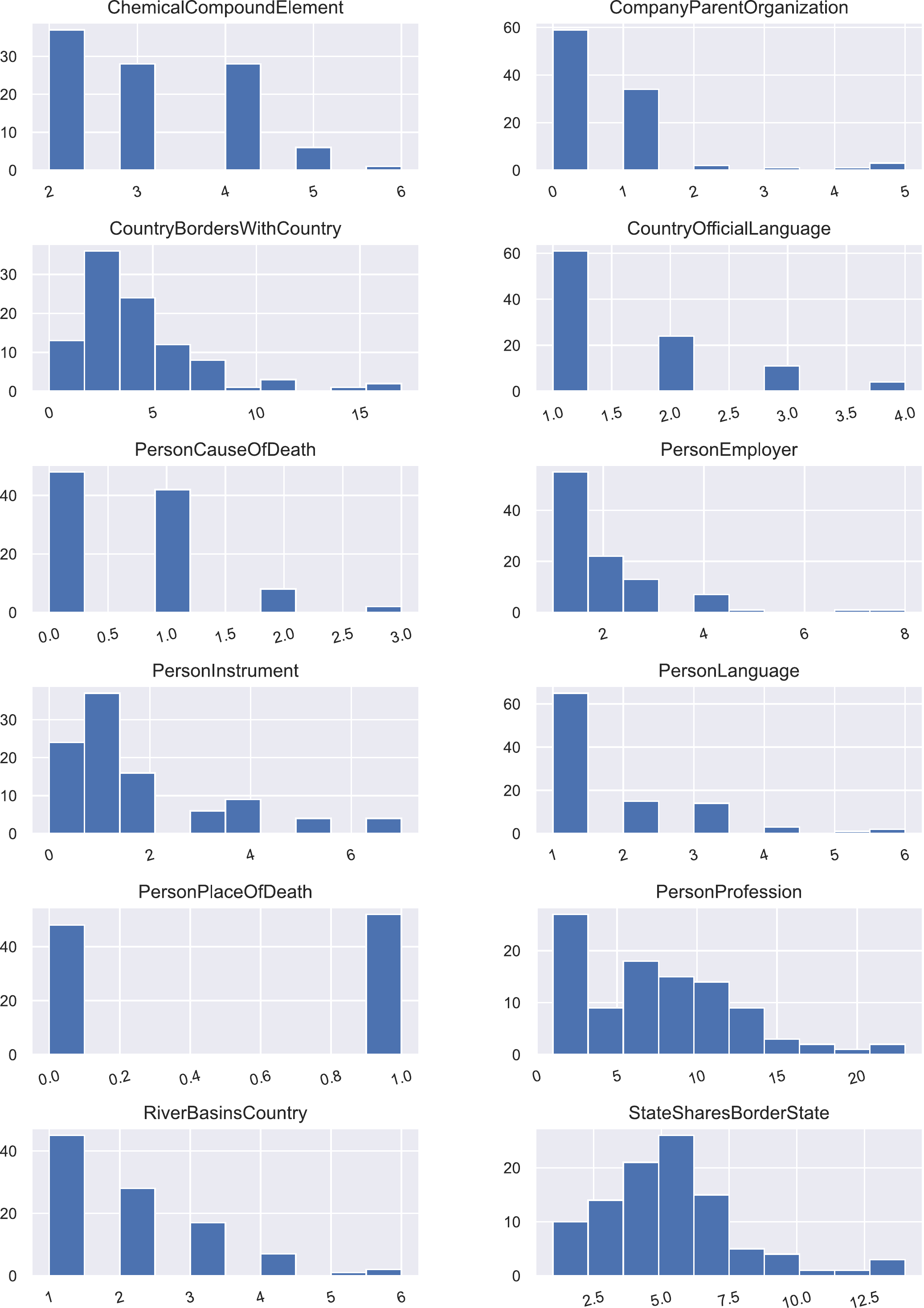}
    \caption{The number of answers per relation type for the training set provided by the organisers.}
    \label{fig:train_set_stats}
\end{figure}

\begin{figure}[ht]
    \centering
    \includegraphics[width=\linewidth]{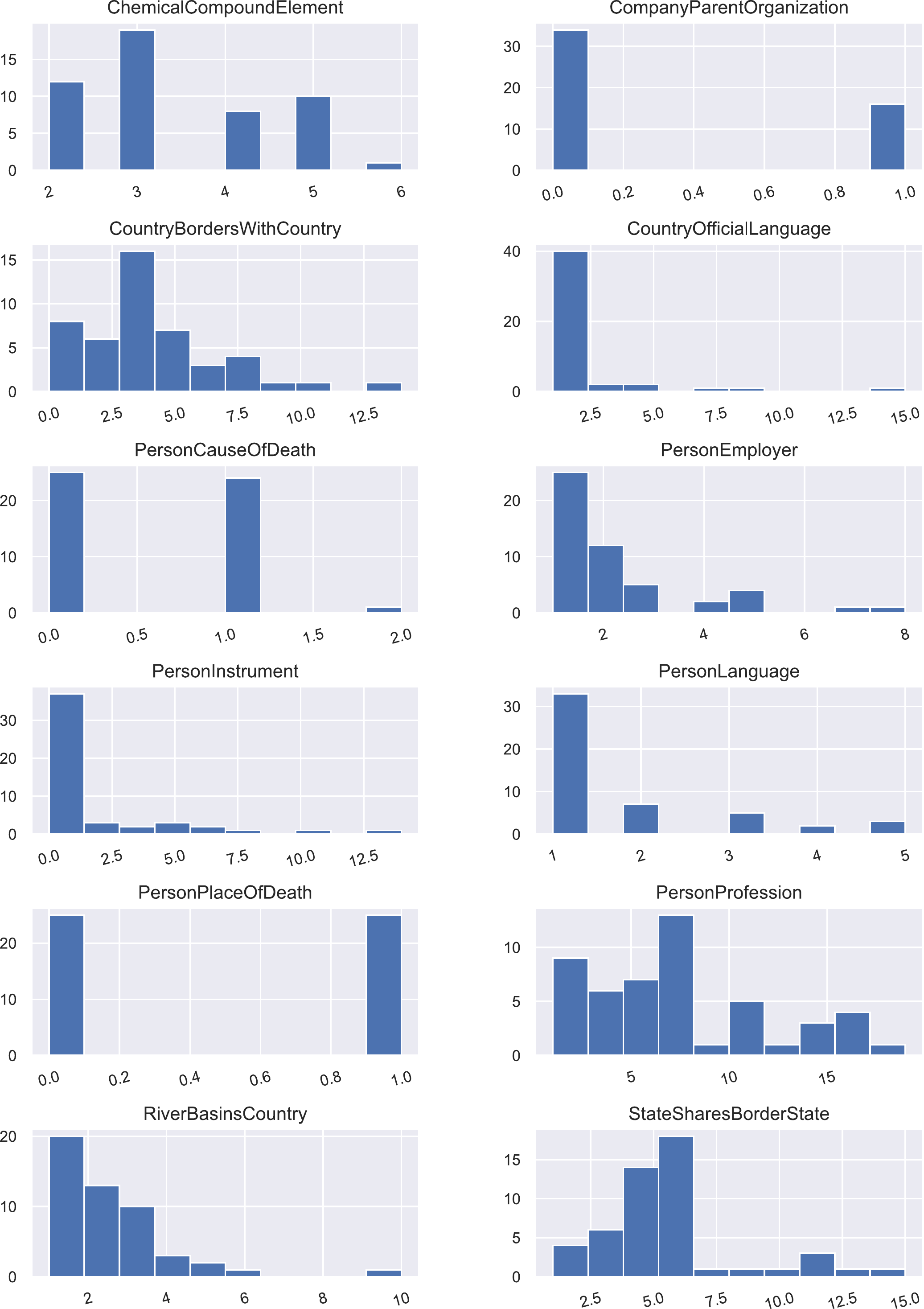}
    \caption{The number of answers per relation type for the development set provided by the organisers.}
    \label{fig:dev_set_stats}
\end{figure}

\begin{table}[ht]
\centering
\caption{The mean and standard deviation (std) of the number of object-entities per relation. Since some object-entities have alternative labels, we also count the alternative labels. The values are rounded to 2 decimal places.}
\label{tab:answer_count_averages}
\scriptsize{
\begin{tabular}{lrr}
\toprule
                          & \multicolumn{2}{c}{Number of Object Entities}                         \\ \cline{2-3} 
                          & \multicolumn{1}{c}{mean} & \multicolumn{1}{c}{std} \\ \midrule
\textbf{Relation Type}                  &                          &                         \\ \midrule
CompanyParentOrganization & 0.32                 & 0.47                \\ 
PersonPlaceOfDeath        & 0.50                 & 0.51                \\ 
PersonCauseOfDeath        & 0.52                 & 0.54                \\ 
PersonLanguage            & 1.70                 & 1.18                \\ 
PersonInstrument          & 1.86                 & 2.81                \\ 
CountryOfficialLanguage   & 2.06                 & 2.47                \\ 
PersonEmployer            & 2.14                 & 1.65                \\ 
RiverBasinsCountry        & 2.28                 & 1.67                \\ 
ChemicalCompoundElement   & 3.38                 & 1.12                \\ 
CountryBordersWithCountry & 4.04                 & 2.69                \\ 
StateSharesBorderState    & 5.62                 & 2.92                \\ 
PersonProfession          & 7.42                 & 4.85                \\ \bottomrule
\end{tabular}
}
\end{table}

\subsubsection{Problems arising from Alternative Labels}
The LM-KBC challenge does not include entity linking. Instead, predicted entities are scored against a list of their aliases in the LM-KBC dataset. However, we noticed these lists are often incomplete. For example, for the "National Aeronautics and Space Administration", the extremely common and widely used abbreviation "NASA", is not included in the list of aliases. Another example occurs when the model predicts Aluminum (US and Canadian English) where the ground truth only has Aluminium (British English; a term globally adopted), a lower score gets obtained. Hence, if the model predicts Aluminum or NASA, the predictions are deemed incorrect.

\subsection{Prompts} \label{prompt-appendix}

Here, we show the templates we used to generate the prompts for the different relations. In our template, we use \verb|{subject_entity}| to refer to the head entity for which we are predicting the tail entities for. The generated prompts were used for the following models: Ada, Babbage, Curie and Davinci.

\subsubsection{CountryBordersWithCountry}

\begin{verbatim}
Which countries neighbour Dominica?
['Venezuela']

Which countries neighbour North Korea?
['South Korea', 'China', 'Russia']

Which countries neighbour Serbia?
['Montenegro', 'Kosovo', 'Bosnia and Herzegovina', 'Hungary', 
'Croatia', 'Bulgaria',  'Macedonia', 'Albania', 'Romania']

Which countries neighbour Fiji?
[]

Which countries neighbour {subject_entity}?
\end{verbatim}

\subsubsection{CountryOfficialLanguage}

\begin{verbatim}
Suriname CountryOfficialLanguage: ['Dutch']

Canada CountryOfficialLanguage: ['English', 'French']

Singapore CountryOfficialLanguage: ['English', 'Malay', 'Mandarin', 
'Tamil']

Sri Lanka CountryOfficialLanguage: ['Sinhala', 'Tamil']

{subject_entity} CountryOfficialLanguage:         
\end{verbatim}

\subsubsection{StateSharesBorderState}

\begin{verbatim}
San Marino StateSharesBorderState: ['San Leo', 'Acquaviva', 
'Borgo Maggiore', 'Chiesanuova', 'Fiorentino']

Whales StateSharesBorderState: ['England']

Liguria StateSharesBorderState: ['Tuscany', 'Auvergne-Rhoone-Alpes', 
'Piedmont', 'Emilia-Romagna']

Mecklenberg-Western Pomerania StateSharesBorderState: ['Brandenburg', 
'Pomeranian', 'Schleswig-Holstein', 'Lower Saxony']

{subject_entity} StateSharesBorderState:
\end{verbatim}

\subsubsection{RiverBasinsCountry}

\begin{verbatim}
Drava RiverBasinsCountry: ['Hungary', 'Italy', 'Austria', 
'Slovenia', 'Croatia']

Huai river RiverBasinsCountry: ['China']

Paraná river RiverBasinsCountry: ['Bolivia', 'Paraguay', 
Argentina', 'Brazil']

Oise RiverBasinsCountry: ['Belgium', 'France']

{subject_entity} RiverBasinsCountry:
\end{verbatim}

\subsubsection{ChemicalCompoundElement}

\begin{verbatim}
Water ChemicalCompoundElement: ['Hydrogen', 'Oxygen']

Bismuth subsalicylate ChemicalCompoundElement: ['Bismuth']

Sodium Bicarbonate ChemicalCompoundElement: ['Hydrogen', 'Oxygen', 
'Sodium', 'Carbon']

Aspirin ChemicalCompoundElement: ['Oxygen', 'Carbon', 'Hydrogen']

{subject_entity} ChemicalCompoundElement:
\end{verbatim}

\subsubsection{PersonLanguage}

\begin{verbatim}
Aamir Khan PersonLanguage: ['Hindi', 'English', 'Urdu']

Pharrell Williams PersonLanguage: ['English']

Xabi Alonso PersonLanguage: ['German', 'Basque', 'Spanish', 'English']

Shakira PersonLanguage: ['Catalan', 'English', 'Portuguese', 'Spanish', 
'Italian', 'French']

{subject_entity} PersonLanguage:
\end{verbatim}

\subsubsection{PersonProfession}

\begin{verbatim}
What is Danny DeVito's profession?
['Comedian', 'Film Director', 'Voice Actor', 'Actor', 'Film Producer', 
'Film Actor', 'Dub Actor', 'Activist', 'Television Actor']

What is David Guetta's profession?
['DJ']

What is Gary Lineker's profession?
['Commentator', 'Association Football Player', 'Journalist', 
'Broadcaster']

What is Gwyneth Paltrow's profession?
['Film Actor','Musician']

What is {subject_entity}'s profession?
\end{verbatim}

\subsubsection{PersonInstrument}

\begin{verbatim}
Liam Gallagher PersonInstrument: ['Maraca', 'Guitar']

Jay Park PersonInstrument: ['None']

Axl Rose PersonInstrument: ['Guitar', 'Piano', 'Pander', 'Bass']

Neil Young PersonInstrument: ['Guitar']

{subject_entity} PersonInstrument:
\end{verbatim}

\subsubsection{PersonEmployer}

\begin{verbatim}
Where is or was Susan Wojcicki employed?
['Google']

Where is or was Steve Wozniak employed?
['Apple Inc', 'Hewlett-Packard', 'University of Technology Sydney', 'Atari']

Where is or was Yukio Hatoyama employed?
['Senshu University','Tokyo Institute of Technology']

Where is or was Yahtzee Croshaw employed?
['PC Gamer', 'Hyper', 'Escapist']

Where is or was {subject_entity} employed?
\end{verbatim}

\subsubsection{PersonPlaceOfDeath}

\begin{verbatim}
What is the place of death of Barack Obama?
[]

What is the place of death of Ennio Morricone?
['Rome']

What is the place of death of Elon Musk?
[]

What is the place of death of Prince?
['Chanhassen']

What is the place of death of {subject_entity}?
\end{verbatim}

\subsubsection{PersonCauseOfDeath}

\begin{verbatim}
André Leon Talley PersonCauseOfDeath: ['Infarction']

Angela Merkel PersonCauseOfDeath: ['None']

Bob Saget PersonCauseOfDeath: ['Injury', 'Blunt Trauma']

Jamal Khashoggi PersonCauseOfDeath: ['Murder']

{subject_entity} PersonCauseOfDeath:
\end{verbatim}

\subsubsection{CompanyParentOrganization}

\begin{verbatim}
Microsoft CompanyParentOrganization: ['None']

Sony CompanyParentOrganization: ['Sony Group']

Saab CompanyParentOrganization: ['Saab Group', 'Saab-Scania', 
'Spyker N.V.', 'National Electric Vehicle Sweden'', 'General Motors']

Max Motors CompanyParentOrganization: ['None]

{subject_entity} CompanyParentOrganization:
\end{verbatim}

\subsection{Failure cases} \label{sec:failure-cases}

Here, we list three failure examples for each relation for the Davinci model. A comprehensive list of failure cases can be found under \href{https://github.com/HEmile/iswc-challenge/tree/main/failure_cases}{https://github.com/HEmile/iswc-challenge/tree/main/failure\_cases}.

\subsubsection{CountryBordersWithCountry}

\begin{verbatim}
SubjectEntity: Bahrain
Ground Truth: ['iran', 'saudi arabia']
GPT-3 Prediction: ['qatar', 'saudi arabia', 'united arab emirates']

SubjectEntity: Barbados
Ground Truth: []
GPT-3 Prediction: ['trinidad and tobago']

SubjectEntity: Cuba
Ground Truth: ['united states of america', 'usa']
GPT-3 Prediction: ['bahamas', 'haiti', 'jamaica', 
'turks and caicos islands', 'united states']
\end{verbatim}

\subsubsection{CountryOfficialLanguage}

\begin{verbatim}
SubjectEntity: Afghanistan
Ground Truth: ['arabic', 'baluchi', 'dari', 'nuristani', 'pamir', 
'pashayi', 'pashto', 'turkmen', 'uzbek']
GPT-3 Prediction: ['dari', 'pashto']

SubjectEntity: Botswana
Ground Truth: ['english']
GPT-3 Prediction: ['setswana']

SubjectEntity: Zimbabwe
Ground Truth: ['barwe', 'chewa', 'english', 'kalanga', 'khoisan', 
'nambya', 'ndau', 'ndebele', 'northern ndebele', 'sesotho', 'shona', 
'tonga', 'tsonga', 'tswana', 'venda', 'xhosa']
GPT-3 Prediction: ['chewa', 'english', 'ndebele', 'shangaan', 'shona', 
'sotho', 'tonga', 'venda']
\end{verbatim}

\subsubsection{StateSharesBorderState}

\begin{verbatim}
SubjectEntity: Andalusia
Ground Truth: ['beja', 'castile-la mancha', 'extremadura', 'faro', 
'gibraltar', 'murcia', 'region of murcia']
GPT-3 Prediction: ['castilla-la mancha', 'ceuta', 'extremadura', 
'melilla', 'murcia']

SubjectEntity: Obwalden
Ground Truth: ['canton of bern', 'canton of lucerne', 'lucerne', 
'nidwalden', 'schwyz', 'uri']
GPT-3 Prediction: ['berne', 'uri', 'valais']

SubjectEntity: Zagreb
Ground Truth: ['krapina-zagorje', 'krapina-zagorje county', 
'zagreb', 'zagreb county']
GPT-3 Prediction: ['bosnia and herzegovina', 'serbia', 'slovenia']
\end{verbatim}

\subsubsection{RiverBasinsCountry}

\begin{verbatim}
SubjectEntity: Didi Liakhvi river
Ground Truth: ['georgia']
GPT-3 Prediction: ['georgia', 'russia']

SubjectEntity: Dniester
Ground Truth: ['moldova', 'ukraine']
GPT-3 Prediction: ['moldova', 'romania', 'ukraine']

SubjectEntity: San
Ground Truth: ['poland', 'ukraine']
GPT-3 Prediction: ['mexico']
\end{verbatim}

\subsubsection{ChemicalCompoundElement}

\begin{verbatim}
SubjectEntity: Aluminum Iodide
Ground Truth: ['aluminium', 'iodine']
GPT-3 Prediction: ['aluminum', 'iodine']

SubjectEntity: Atracurium Besylate
Ground Truth: ['carbon', 'hydrogen', 'nitrogen', 'oxygen', 'sulfur']
GPT-3 Prediction: ['atracurium', 'hydrogen', 'oxygen']

SubjectEntity: Flavin Adenine Dinucleotide
Ground Truth: ['carbon', 'hydrogen', 'nitrogen', 'oxygen', 'phosphorus']
GPT-3 Prediction: ['adenine', 'dinucleotide', 'flavin']
\end{verbatim}

\subsubsection{PersonLanguage}

\begin{verbatim}
SubjectEntity: A. R. Rahman
Ground Truth: ['hindi', 'tamil', 'telugu']
GPT-3 Prediction: ['bengali', 'hindi', 'kannada', 'malayalam', 
'marathi', 'tamil', 'telugu']

SubjectEntity: Akon
Ground Truth: ['english']
GPT-3 Prediction: ['english', 'french', 'wolof']

SubjectEntity: Genelia D'Souza
Ground Truth: ['hindi', 'malayalam', 'tamil', 'telugu']
GPT-3 Prediction: ['bengali', 'english', 'hindi', 'marathi', 'telugu']
\end{verbatim}

\subsubsection{PersonProfession}

\begin{verbatim}
SubjectEntity: Elon Musk
Ground Truth: ['engineer', 'entrepreneur', 'inventor', 'programmer']
GPT-3 Prediction: ['businessperson', 'engineer', 'inventor', 
'investor', 'philanthropist', 'product designer']

SubjectEntity: J. K. Rowling
Ground Truth: ['author', "children's writer", 'executive producer', 
'film producer', 'novelist', 'producer', 'producer', 'screenwriter', 
'writer', 'writer']
GPT-3 Prediction: ['author', 'film producer', 'philanthropist', 
'screenwriter']

SubjectEntity: Lady Gaga
Ground Truth: ['singer', 'songwriter']
GPT-3 Prediction: ['actress', 'dancer', 'philanthropist', 
'record producer', 'singer', 'songwriter']
\end{verbatim}

\subsubsection{PersonInstrument}

\begin{verbatim}
SubjectEntity: Ronan Keating
Ground Truth: []
GPT-3 Prediction: ['piano']

SubjectEntity: Taeyang
Ground Truth: []
GPT-3 Prediction: ['guitar']

SubjectEntity: Chester Bennington
Ground Truth: []
GPT-3 Prediction: ['guitar', 'piano']
\end{verbatim}

\subsubsection{PersonEmployer}

\begin{verbatim}
SubjectEntity: Kent Beck
Ground Truth: ['meta platforms']
GPT-3 Prediction: ['facebook', 'three rivers institute']

SubjectEntity: Serena Williams
Ground Truth: ['unicef']
GPT-3 Prediction: ['tennis']

SubjectEntity: Guido van Rossum
Ground Truth: ['microsoft']
GPT-3 Prediction: ['dropbox', 'google']
\end{verbatim}

\subsubsection{PersonPlaceOfDeath}

\begin{verbatim}SubjectEntity: Avicii
Ground Truth: ['muscat']
GPT-3 Prediction: ['muscat, oman']

SubjectEntity: John Coltrane
Ground Truth: ['huntington']
GPT-3 Prediction: ['new york city']

SubjectEntity: Rachel Caine
Ground Truth: ['texas']
GPT-3 Prediction: ['']
\end{verbatim}

\subsubsection{PersonCauseOfDeath}

\begin{verbatim}
SubjectEntity: Ahmed Zewail
Ground Truth: ['lymphoma', 'spinal cord lymphoma']
GPT-3 Prediction: ['cancer']

SubjectEntity: Avicii
Ground Truth: ['exsanguination']
GPT-3 Prediction: ['suicide']

SubjectEntity: Ennio Morricone
Ground Truth: ['femoral fracture', 'fracture']
GPT-3 Prediction: [''] 
\end{verbatim}

\subsubsection{CompanyParentOrganization}

\begin{verbatim}
SubjectEntity: Aston Martin lagonda
Ground Truth: []
GPT-3 Prediction: ['aston martin lagonda global holdings plc']

SubjectEntity: Austro-Daimler
Ground Truth: []
GPT-3 Prediction: ['daimler ag']

SubjectEntity: Hyundai Motor Company
Ground Truth: ['hyundai']
GPT-3 Prediction: ['hyundai motor group']
\end{verbatim}

\subsection{Language Model Size}

Table \ref{table:results-scaling} shows the values of the scaling experiments. These values were used to produce Figure \ref{fig:performance_per_model_type}.

\begin{table*}[h]
\centering
\scriptsize{
\caption{
\label{table:results-scaling}Precision, Recall and F1-score for the GPT-3 models with varying embedding dimensions. Best scores are in \textbf{bold}.}
\begin{tabular}{|p{5cm}|p{2cm}|p{2cm}|p{2cm}|}
\hline
\textbf{Method} & \textbf{Precision} & \textbf{Recall} & \textbf{F1-score} \\ \hline \hline
Baseline (BERT) & 0.175 & 0.129 & 0.140 \\ \hline
Ada  & 0.180  & 0.194 & 0.161 \\ \hline
Babbage  & 0.325   & 0.263 & 0.269 \\ \hline
Curie  & 0.378 & 0.375 & 0.343 \\ \hline
Davinci  & \textbf{0.707} & \textbf{0.694} & \textbf{0.677} \\ \hline
\end{tabular}}
\end{table*}

\end{document}